\documentclass[10pt,twocolumn,letterpaper]{article}
\usepackage[T1]{fontenc}
\usepackage[datasets,pagenumbers]{taylorgeospatial}

\usepackage{microtype}
\usepackage{graphicx}
\usepackage{tikz}
\usepackage{amsmath,amssymb}
\usepackage{booktabs}
\usepackage{multirow}
\usepackage{makecell}
\usepackage{caption}
\usepackage{siunitx}
\renewcommand{\arraystretch}{0.92}
\usepackage{xcolor}
\usepackage{colortbl}
\usepackage[pagebackref,breaklinks,colorlinks]{hyperref}
\usepackage[capitalize,noabbrev]{cleveref}

\definecolor{CiteGray}{HTML}{3F51B5}
\definecolor{FigPink}{HTML}{FF4081}
\hypersetup{
    colorlinks=true,
    citecolor=CiteGray, %
    linkcolor=FigPink,  %
    urlcolor=CiteGray   %
}

\definecolor{earthDark}{rgb}{0.30, 0.40, 0.18}
\definecolor{earthSienna}{rgb}{0.55, 0.34, 0.20}
\definecolor{recred}{HTML}{FF4F2C}

\def\wacvPaperID{3617}
\def\confName{WACV}
\def\confYear{2027}

\begin{document}

\title{Fields of the Planet: Field Boundary Mapping Beyond 10\,m}
\author{Isaac Corley\textsuperscript{1} \quad Caleb Robinson\textsuperscript{2} \quad Hannah Kerner\textsuperscript{1,3} \quad Jennifer Marcus\textsuperscript{1}\\[2pt]
\textsuperscript{1}Taylor Geospatial \quad \textsuperscript{2}Microsoft AI for Good Research Lab \quad \textsuperscript{3}Arizona State University}

\twocolumn[{%
\maketitle
\centering
\vspace{1pt}
{\normalsize\href{https://research.taylorgeospatial.org/fields-of-the-planet}{research.taylorgeospatial.org/fields-of-the-planet}\par}
\vspace{11pt}
\includegraphics[width=0.99\textwidth]{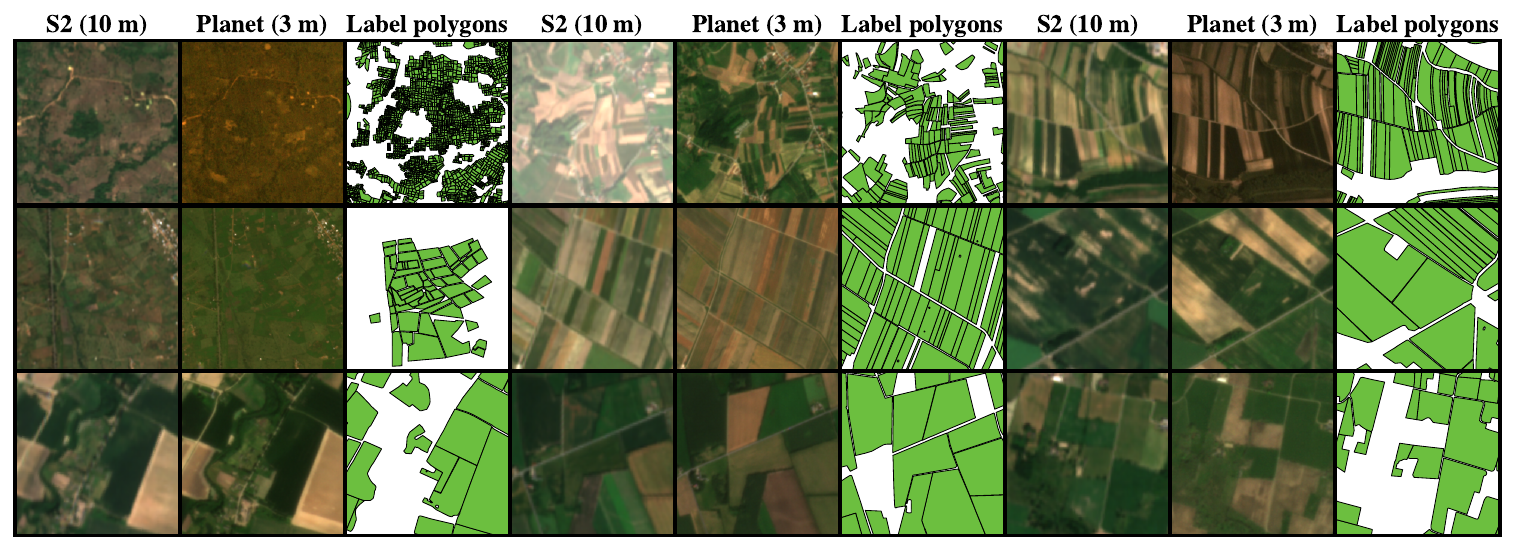}
\captionof{figure}{\textbf{Fields of the Planet (FTP) pairs Fields of The World (FTW) Sentinel-2 patches with co-registered 3\,m PlanetScope imagery and field-boundary labels.} Each triplet shows the FTW Sentinel-2 patch at 10\,m, the paired PlanetScope surface-reflectance (SR) image at 3\,m (4-band; RGB shown), and the ground-truth field-boundary labels from the original FTW vector polygons.}\label{fig:hero}
\vspace{15pt}
}]
\begin{abstract}
Field-boundary accuracy depends on the model, sensor resolution, label discretization, and evaluation procedure. Measuring the contribution of sensor resolution therefore requires a controlled comparison across sensors. We present \textbf{Fields of the Planet (FTP)}, a 3\,m PlanetScope companion to Fields of The World (FTW) containing 133{,}168 co-registered patch-window targets across 24 countries. FTP preserves FTW's vector polygons, seasonal windows, and dataset splits, enabling direct comparison with 10\,m Sentinel-2. We first quantify the resolution ceiling by rasterizing and vectorizing the same ground-truth parcels at each resolution to determine which fields the label grid can represent. We evaluate predicted parcels with polygon-level and boundary metrics. With identical PRUE+ training and post-processing across sensors, plus an upsampled Sentinel-2 control, we isolate gains from image content. PlanetScope raises panoptic quality (PQ) from 21.0 to 35.5, improves PQ on sub-$0.5$\,ha fields from 5.8 to 15.7, and reduces matched-boundary error from 18.6\,m to 7.4\,m. Pixel-level metrics understate this improvement, so field-boundary benchmarks should account for sensor resolution.
\end{abstract}

\section{Introduction}

Field boundaries are inputs to crop-yield estimation, irrigation accounting, subsidy verification, and global food-security monitoring~\cite{beckerreshef2020geoglam,defourny2019sen2agri}. Farms under 2\,ha number roughly 510 million and account for about 84\% of the world's farms \emph{by count}~\cite{lowder2021farms,lowder2016farms}, yet their parcels are often only tens of meters across. Irrigation and yield are aggregated within fields, and subsidy verification checks declared areas per parcel~\cite{beckerreshef2020geoglam,defourny2019sen2agri}. Downstream systems use parcel geometries rather than pixel masks. A prediction that merges neighboring fields can fail its intended use even when its intersection-over-union (IoU) is high; therefore, field-boundary benchmarks should measure whether the sensor and model recover usable parcel geometry.

\emph{Fields of The World} (FTW)~\cite{kerner2024ftw} provides over 1.62\,M parcel polygons across 24 countries and 25 labeled regions, paired with 10\,m Sentinel-2 imagery~\cite{drusch2012sentinel2}. However, prior studies report that Sentinel-2 struggles to delineate fields only a few tens of meters across~\cite{garciapedrero2017boundary,persello2019delineation,wang2022unlocking}. Rasterizing FTW's vector polygons onto a 10\,m grid merges many adjacent parcels into single connected components, and only $\sim$$7.5\%$ of sub-$0.5$\,ha fields remain separable at 10\,m, compared with $\sim$$88\%$ at 3\,m (\Cref{tab:representation_ceiling}). These losses occur during label rasterization, before a model makes any prediction.

Pixel overlap metrics like IoU reward covered area but do not directly count merged, missed, or hallucinated parcels. To assess whether predictions form usable field geometries, we instead evaluate how accurately individual field polygons and their boundaries are identified. Predictions are vectorized, matched to the original ground-truth polygons, and scored with object F1, panoptic quality (PQ), size-stratified PQ, and meter-scale boundary error. Boundary error is computed only for matched polygons, whereas missed and hallucinated fields affect recognition quality and object F1. We then study whether gains in pixel- and polygon-level metrics for field-boundary mapping are correlated.

We introduce \textbf{Fields of the Planet (FTP)}, which pairs 3\,m PlanetScope surface-reflectance imagery~\cite{planet2017} with FTW's polygons, seasonal windows, and tile-level splits (\Cref{fig:hero}). The 70{,}484 FTW patches define 140{,}968 planting- and harvest-window targets. FTP pairs 133{,}168 of them (94.5\%) with cloud-screened PlanetScope acquisitions across the same 24 countries. With this pairing, annotations, geographic regions, seasonal target definitions, split assignments, architecture, training recipe, and post-processing remain fixed while the imagery source changes.

\paragraph{Contributions.}
Our contributions are: (i) \emph{Fields of the Planet}, a 3\,m companion to FTW that preserves its vector labels, seasonal windows, and train/test splits across 24 countries (\Cref{sec:dataset}); (ii) a polygon-level evaluation that measures parcel detection, shape, field-size effects, and matched-boundary localization separately (\Cref{sec:eval-metrics}); (iii) rasterization and upsampling controls that isolate the effect of label and output resolution while holding other experimental settings fixed (\Cref{tab:representation_ceiling,tab:resolution_ablation}); and (iv) PRUE+ baselines trained with one recipe for both sensors, evaluation of released instance-segmentation models~\cite{lavreniuk2025delineate}, and the code and quality metadata needed to reproduce the comparison. Our experiments find that PlanetScope improves polygon recovery and matched-boundary localization most on fields smaller than $0.5$\,ha.

\section{Related work}

\paragraph{Field-boundary benchmarks.}
Broad geographic coverage and imagery fine enough to distinguish smallholder parcels remain difficult to obtain in the same benchmark. AI4SmallFarms~\cite{persello2023ai4smallfarms} releases roughly 50{,}000 annotated smallholder parcels across South and Southeast Asia with 10\,m Sentinel-2 imagery. PASTIS~\cite{garnot2021pastis} and PASTIS-HD~\cite{pastishd} provide panoptic agricultural parcel annotations for France using Sentinel-2 time series and aligned high-resolution SPOT 6/7 imagery. High-resolution resources also include PlanetScope benchmarks for France and India~\cite{wang2022unlocking} and Africa-wide labels digitized from NICFI Planet basemaps~\cite{estes2024africa}. FTW~\cite{kerner2024ftw} instead provides semantic, boundary, and instance annotations across 24 countries. FTP preserves this geographic coverage, the original vector annotations, and dataset splits while adding paired 3\,m imagery.

\paragraph{PlanetScope for field-boundary delineation.}
Wang et al.~\cite{wang2022unlocking} use 3\,m PlanetScope imagery for parcel delineation in France and India. Khallaghi et al.~\cite{khallaghi2025generalization} and Estes et al.~\cite{estes2024africa} use Planet basemap mosaics for cropland or field-boundary labeling. These mosaics improve temporal coverage but differ from temporally matched, scene-level imagery. DECODE~\cite{waldner2021decode} trains PlanetScope field-boundary models over Australia, but its data and models were never released. Delineate Anything~\cite{lavreniuk2025delineate} treats delineation as resolution-agnostic instance segmentation using a multi-sensor RGB corpus concentrated in Europe. Its successor, Delineate Anything v2~\cite{lavreniuk2026delineatev2}, scales the training corpus to 73\,M instances across 61 countries. Despite this broad geographic coverage, the released Delineate Anything models achieve substantially lower polygon-level accuracy than our PRUE+ baselines in our experiments, even after tuning inference resolution and confidence thresholds (\Cref{tab:polygon_metrics}).

\paragraph{Parcel-level evaluation.}
Field-boundary benchmarks commonly begin with raster masks, even when the final product is a polygon map~\cite{kerner2024ftw,muhawenayo2026prue,persello2019delineation,wang2022unlocking,waldner2021decode}. The canonical PASTIS benchmark uses satellite time-series inputs from France alone and represents agricultural parcels as panoptic instances~\cite{garnot2021pastis}, following the broader decomposition of panoptic quality into recognition and segmentation quality~\cite{kirillov2019panoptic}. General vision benchmarks also use stricter overlap thresholds and boundary-aware measures to separate detection from localization quality~\cite{cheng2021biou,lin2014coco}. FTP evaluates parcel detection and shape across field sizes by matching predicted polygons to the original ground-truth vectors. Boundary distances are measured in meters for matched pairs, while missed and hallucinated fields count as recognition errors. We compare these polygon-level results with FTW's raster metrics to test whether pixel-level gains translate into better field boundaries.

\section{Dataset}
\label{sec:dataset}
We construct FTP by selecting seasonally matched PlanetScope scenes and screening image quality at the patch level before generating the raster labels and release index.

\paragraph{Seasonal pairing.}
For each FTW patch and each of its two seasonal Sentinel-2 timestamps, we query the Planet Data API for PlanetScope scenes using a 10\% scene-level cloud-cover threshold. We select the scene acquired closest to the target timestamp that covers the patch bounds, then download and clip its \textit{ortho\_analytic\_4b\_sr} product (4-band surface reflectance, $\sim$3\,m, uint16 GeoTIFF) and the quality flags contained in the Usable Data Mask (UDM2). We retain a patch only when both its planting and harvest windows are successfully extracted. Otherwise, we drop the paired sample. FTW imagery is reprojected to EPSG:4326, whereas each FTP patch retains its local UTM coordinate reference system (CRS). FTP image dimensions follow the patch footprint (typically $\sim$510$\times$330\,px) rather than FTW's fixed 256$\times$256 image size.

\paragraph{Patch-level quality control.}
Scene-level cloud cover is too coarse for patch-level dataset construction. A scene can pass the 10\% filter while clouds, haze, or shadows overlap a particular FTW patch. We compute per-class coverage directly over every clipped UDM2 mask. Among the 129{,}490 tiles with valid UDM2 statistics, 89.7\% satisfy the strict usable-tile criterion (clear $\ge 95\%$, unusable $\le 5\%$); \Cref{tab:udm2} reports the full coverage breakdown.

For every window that fails a UDM2 threshold, we search for alternative PlanetScope scenes from the same season ($\pm$60 days), evaluate UDM2 coverage over the patch footprint, and replace the image and mask only when an alternative strictly improves quality. This step handles scenes that are mostly clear but cloudy over the patch footprint. Usable-tile rates range from 48\% in Portugal to over 99\% in Brazil and Croatia (\Cref{app:udm2_country}). The released Parquet index records scene identifiers, target and acquisition dates, scene coverage, and per-class UDM2 statistics for each window. We also release the resampling tool and QA table so users can audit each final pairing or apply stricter filters.

\paragraph{Raster labels.}
\label{sec:labels}
We download the published FTW polygons (1{,}627{,}378 across all regions)~\cite{ftw-source-coop}, clip them to each PlanetScope patch's bounds, reproject them to its CRS, and rasterize them using the image's height, width, and transform. The resulting mask has three classes: background, field interior, and field boundary. We draw the boundary class by buffering each polygon's exterior ring by $3$\,m ($\approx$1 pixel per side), which keeps touching parcels separate.

\paragraph{Label provenance.}
Using FTW labels on a $3\times$ finer grid assumes that the source geometries carry precision beyond the 10\,m Sentinel-2 raster. FTW meets this requirement because its native vector geometries come from government cadastral and declared-parcel registries and regional labeling campaigns~\cite{kerner2024ftw}, not annotations drawn on the 10\,m images. Rasterizing these vectors at $3$\,m preserves more of their existing geometric detail.

\paragraph{Rasterization ceiling.}
Rasterization merges some field instances before model training. We isolate this representational limit by rasterizing the source FTW polygons at each sensor's native resolution, vectorizing the masks, and matching the recovered polygons back to the source geometries (\Cref{tab:representation_ceiling}). At $10$\,m, only $7.5\%$ of sub-$0.5$\,ha parcels remain recoverable as separate polygons, compared with $88.3\%$ at $3$\,m. Recovery of medium fields also increases, from $73.6\%$ to $99.2\%$, while large-field recovery changes little ($96.6\%$ to $98.1\%$). Adjacent parcels remain separate only when the raster can place a boundary pixel between them. At $10$\,m, this boundary can be wider than the gap between small parcels, merging neighboring fields before training begins. The ceiling is independent of any model and upper-bounds parcel recovery through the raster-label pipeline.

\begin{table}[h!]
    \centering
    \caption{\textbf{Rasterization ceiling.} Percentage of source FTW polygons recovered as separate parcels after rasterizing the labels at each resolution and vectorizing them back. At $10$\,m, only $7.5\%$ of sub-$0.5$\,ha parcels remain recoverable, compared with $88.3\%$ at $3$\,m.}
    \label{tab:representation_ceiling}
    \setlength{\tabcolsep}{6pt}
\begin{tabular}{@{}lrcc@{}}
\toprule
GT field size & $n$ & 3\,m & 10\,m \\
\midrule
Small ($<0.5$\,ha) & 19{,}135 & \textbf{88.3} & 7.5 \\
Medium ($0.5$--$2$\,ha) & 7{,}793 & 99.2 & 73.6 \\
Large ($\ge2$\,ha) & 2{,}221 & 98.1 & 96.6 \\
\bottomrule
\end{tabular}

\end{table}

\begin{table*}[t!]
    \centering
    \caption{\textbf{Polygon-level results averaged over the ten dense-label held-out countries dominated by smallholder fields.} With the same backbone, FTP (PlanetScope) improves all polygon-level field-boundary metrics (PQ, SQ, RQ, F1, and boundary error) over Sentinel-2. Boundary error is computed on matched polygons only, while missed or hallucinated fields affect RQ/F1; $|\Delta N|/N$ is the normalized per-country polygon-count error. Delineate Anything count errors use per-patch aggregation and are not directly comparable. $^{*}$Released models evaluated without training on FTW or FTP, each at its best swept inference resolution and confidence setting. $^{\dagger}$Pixel IoU is not comparable across sensors due to differences in resolution. $^{\ddagger}$PQ for small ($<0.5$\,ha), medium ($0.5$--$2$\,ha), and large ($\ge2$\,ha) ground-truth fields.}
    \label{tab:polygon_metrics}
    \footnotesize
\setlength{\tabcolsep}{3pt}
\begin{tabular}{@{}l l l ccc c c cc c ccc@{}}
\toprule
 & & & \multicolumn{3}{c}{Panoptic} & & & \multicolumn{2}{c}{\makecell{Bd.\ err\ (m)\,$\downarrow$}} & & \multicolumn{3}{c}{\makecell{PQ by field area$^{\ddagger}$}} \\
\cmidrule(lr){4-6} \cmidrule(lr){9-10} \cmidrule(lr){12-14}
Method & Sensor & Backbone & PQ & SQ & RQ$_{.5}$ & F1$_{[.5{:}.95]}$ & \makecell{$|\Delta N|/N$\,$\downarrow$} & mean & p95 & \makecell{Pixel\\IoU$^{\dagger}$} & PQ$_\mathrm{s}$ & PQ$_\mathrm{m}$ & PQ$_\mathrm{l}$ \\
\midrule
Delineate Anything$^{*}$~\cite{lavreniuk2025delineate} & Planet & YOLO11x & 9.5 & 73.3 & 12.7 & 7.0 & 0.75 & 13.7 & 37.8 & 51.1 & 1.7 & 7.1 & 16.3 \\
Delineate Anything-S$^{*}$~\cite{lavreniuk2025delineate} & Planet & YOLO11n & 3.5 & 70.8 & 4.8 & 2.5 & 0.82 & 13.2 & 34.2 & 40.7 & 0.8 & 2.8 & 6.7 \\
Delineate Anything v2$^{*}$~\cite{lavreniuk2026delineatev2} & Planet & YOLO11x & 7.5 & 75.0 & 10.0 & 5.6 & 0.82 & 9.4 & 25.5 & 27.1 & 4.4 & 11.4 & 14.8 \\
\midrule
FTW-PRUE+~\cite{muhawenayo2026prue} & S2 & EfficientNet-B3 & 21.0 & 71.4 & 28.9 & 14.6 & 0.33 & 18.6 & 54.7 & 61.8 & 5.8 & 25.3 & 33.8 \\
FTW-PRUE+~\cite{muhawenayo2026prue} & S2 & EfficientNet-B7 & 24.2 & 71.0 & 32.8 & 17.2 & 0.35 & 14.4 & 43.4 & 63.6 & 7.5 & 28.4 & 37.7 \\
\midrule
\textbf{FTP-PRUE+ (ours)} & \textbf{Planet} & \textbf{EfficientNet-B3} & \textbf{35.5} & \textbf{75.7} & \textbf{46.2} & \textbf{27.1} & 0.33 & \textbf{7.4} & \textbf{22.8} & 68.8 & \textbf{15.7} & 39.2 & \textbf{52.0} \\
\textbf{FTP-PRUE+ (ours)} & \textbf{Planet} & \textbf{EfficientNet-B7} & 35.4 & 74.4 & 46.1 & 27.0 & \textbf{0.30} & \textbf{7.4} & \textbf{22.8} & \textbf{74.2} & 15.6 & \textbf{40.6} & 50.9 \\
\bottomrule
\end{tabular}

\end{table*}

\paragraph{Scope and licenses.}
FTP covers 66{,}584 of the 70{,}484 FTW patches, or 133{,}168 patch-window targets, drawn from 6{,}113 unique PlanetScope scenes. The constructed imagery dataset totals 102\,GB, with a full breakdown in \Cref{tab:scope} (appendix); the non-random coverage gaps are summarized in \Cref{sec:limitations}. The PlanetScope imagery was acquired under Planet's Education and Research program. Due to licensing restrictions, the public release does not redistribute the imagery. Instead, we provide the original PlanetScope item identifiers and timestamps in the dataset index, together with code to download the imagery through Planet. Reconstructing the imagery dataset requires appropriate Planet access and remains subject to the applicable Planet license terms. The dataset metadata and trained models (\href{https://hf.co/fieldsoftheworld/ftp-b3}{FTP-B3} and \href{https://hf.co/fieldsoftheworld/ftp-b7}{FTP-B7}) are released on Hugging Face under CC-BY-NC 4.0. Labels also carry their per-region FTW licenses, which range from CC-BY to CC-BY-NC by country. We release the \href{https://github.com/taylor-geospatial/fields-of-the-planet}{training and evaluation code} under an MIT license so the benchmark can be reproduced.

\section{Building a PlanetScope baseline}
\label{sec:benchmark}

We use PRUE~\cite{muhawenayo2026prue}, which reports the highest performance among released FTW training recipes, based on a published architecture and training sweep. PRUE+ adapts this baseline to 3\,m imagery through the augmentation and instance-extraction choices described below. For the primary sensor comparison, we hold the U-Net architecture~\cite{ronneberger2015unet}, EfficientNet backbone~\cite{tan2019efficientnet}, loss, data splits, training schedule, checkpoint rule, and post-processing fixed, changing only the imagery source and its native raster grid. We report EfficientNet-B3 and B7 results for both FTW and FTP and evaluate the released Delineate Anything models~\cite{lavreniuk2025delineate} under the same polygon evaluation procedure (\Cref{tab:polygon_metrics}). In \Cref{sec:analysis}, we test how the conclusion changes with the metric, output grid, region, and field size.

\begin{figure}[t!]
    \centering
    \includegraphics[width=\linewidth]{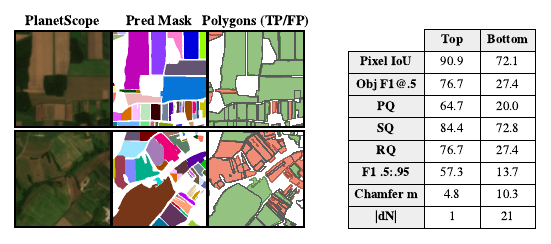}
    \caption{\textbf{Pixel overlap and polygon recovery.}
    Two PlanetScope examples show a successful parcel reconstruction (top) and a failure with high pixel IoU but low object F1 (bottom). Columns show the input, FTP-PRUE+ mask, and vectorized predictions matched to ground truth; green and red denote matched and unmatched predicted polygons. In the bottom row, the mask covers much of the field area, but fragmented parcels yield low object F1.}
    \label{fig:metric_example}
\end{figure}

\subsection{Evaluation metrics}
\label{sec:eval-metrics}
Merging neighboring fields into a single polygon can yield high pixel IoU despite failing to delineate individual parcels. We report pixel IoU to be consistent with PRUE. For the primary evaluation, we vectorize each model's native-grid field-interior mask and compare the resulting parcel objects with the ground-truth FTW vector polygons. We score parcel detection separately from shape and boundary accuracy.

\paragraph{Polygon-level matching.}
We match a predicted polygon to a ground-truth field when their IoU exceeds 0.5. All polygon metrics use the resulting true positives, false positives, and false negatives, which count merged, missed, and hallucinated fields as object errors rather than averaging them into an aggregate pixel score.

\begin{figure*}[t!]
    \centering
    \includegraphics[width=0.95\linewidth]{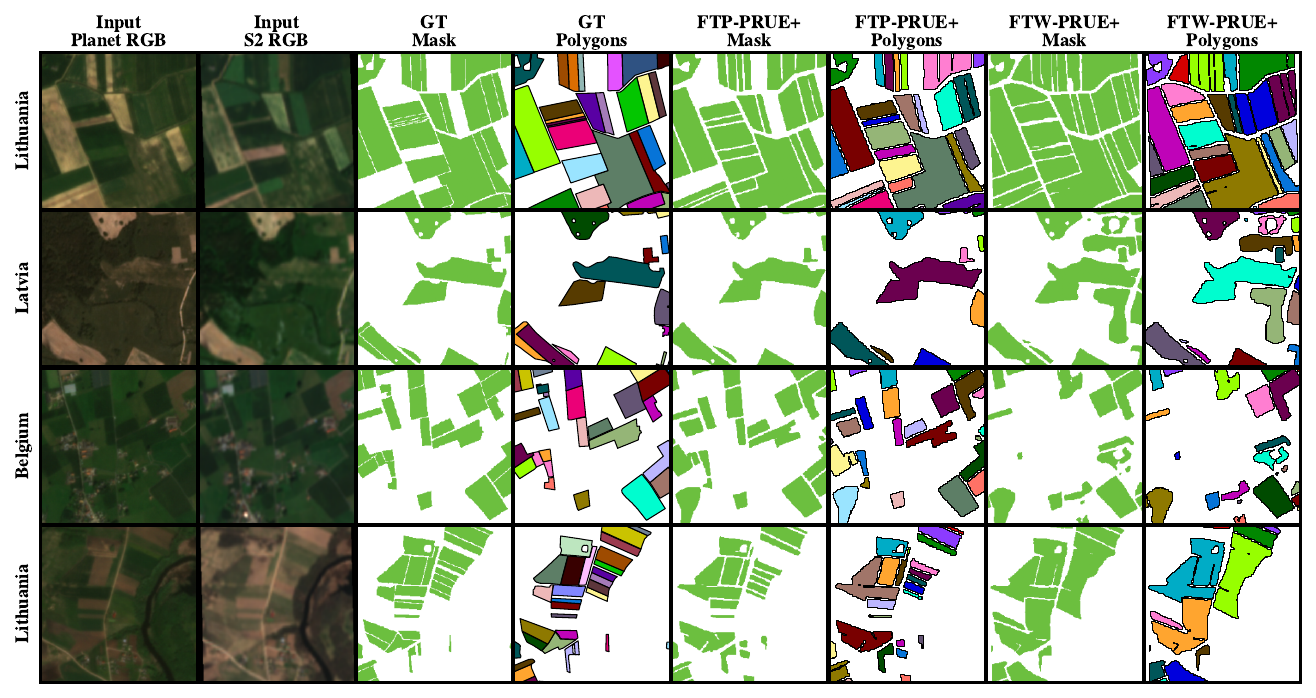}
    \caption{\textbf{Selected cross-sensor parcel reconstructions.} Held-out examples from Belgium, Latvia, and Lithuania contain dense sub-$0.5$\,ha parcels that are difficult to separate at 10\,m resolution. Columns show the 3\,m PlanetScope image, the 10\,m Sentinel-2 image, ground truth, and each model output as a mask/vectorized-polygon pair. FTP-PRUE+ recovers more parcel structure in these examples, whereas the same-recipe Sentinel-2 baseline merges neighboring fields or leaves gaps.}
    \label{fig:qualitative}
\end{figure*}

\paragraph{Object F1 and panoptic quality.}
\textit{Object F1} summarizes the precision and recall of polygon matching. At $\mathrm{IoU}=0.5$, it is equivalent to \textit{recognition quality (RQ)}. \textit{Panoptic quality (PQ)}~\cite{kirillov2019panoptic} multiplies RQ by \textit{segmentation quality (SQ)}, the mean IoU of matched parcels. RQ measures whether fields are found, SQ measures the shape of matched fields, and PQ combines both effects.

\paragraph{Stricter localization metrics.}
A match at $\mathrm{IoU}=0.5$ may still be too coarse for downstream use. We also report object F1 averaged over thresholds $\{0.5,0.55,\ldots,0.95\}$~\cite{lin2014coco}, which requires predictions to remain matched under stricter overlap criteria. Mean symmetric boundary Chamfer distance~\cite{cheng2021biou} reports the distance between predicted and ground-truth boundaries in meters, independent of pixel units. Chamfer is computed \emph{only} on matched polygon pairs; misses and hallucinations affect RQ and F1. \Cref{app:metrics} gives the formal definitions, and \Cref{fig:metric_example} shows why pixel IoU alone is insufficient.

\begin{figure}[t!]
    \centering
    \begingroup
\definecolor{augCoral}{HTML}{FF4F2C}
\definecolor{augBlue}{HTML}{4A78B5}
\definecolor{augNavy}{HTML}{24496D}
\definecolor{augGray}{HTML}{D8D3C6}
\begin{tikzpicture}[x=2.65pt,y=1pt,
  font=\fontfamily{ptm}\fontsize{7.5}{9}\selectfont,
  every node/.style={inner sep=0pt},line cap=butt]
\path[use as bounding box] (-36,-40) rectangle (53,91);
\foreach \x in {0,10,20,30,40,50} {
  \node[anchor=north,font=\fontsize{7}{8}\selectfont] at (\x,-5) {\x};
}
\draw[black!65,line width=0.5pt] (0,0) -- (50,0);
\draw[augGray,line width=0.4pt] (-36,22) -- (50,22);
\fill[augCoral!45!white] (0,67) rectangle (29.0715600,77);
\node[anchor=east] at (-2,72) {PRUE B3 (default)};
\node[anchor=east,fill=white,inner sep=1pt] at (53,72) {29.1};
\fill[augCoral!60!white] (0,52) rectangle (31.9130100,62);
\node[anchor=east] at (-2,57) {+ Preprocessing / Resize};
\node[anchor=east,fill=white,inner sep=1pt] at (53,57) {31.9};
\fill[augCoral!75!white] (0,37) rectangle (32.9371700,47);
\node[anchor=east] at (-2,42) {+ Seasonal Swap / Gamma};
\node[anchor=east,fill=white,inner sep=1pt] at (53,42) {32.9};
\fill[augCoral!90!white] (0,22) rectangle (37.3911600,32);
\node[anchor=east] at (-2,27) {+ Geometry / Noise};
\node[anchor=east,fill=white,inner sep=1pt] at (53,27) {37.4};
\fill[augCoral!100!white] (0,5) rectangle (45.1628800,15);
\node[anchor=east] at (-2,10) {+ All Countries};
\node[anchor=east,fill=white,inner sep=1pt] at (53,10) {45.2};
\draw[augBlue,dash pattern=on 3pt off 2pt,line width=0.65pt] (34.8746200,0) -- (34.8746200,79);
\draw[augNavy,densely dotted,line width=0.65pt] (39.4216300,0) -- (39.4216300,79);
\node at (25,-20) {Object F1 (\%)};
\node at (8.5,-35) {
  \tikz[baseline=-0.5ex] \draw[augBlue,dash pattern=on 3pt off 2pt,line width=0.65pt] (0,0)--(9pt,0);
  Sentinel-2 B3 (34.9)\hspace{8pt}
  \tikz[baseline=-0.5ex] \draw[augNavy,densely dotted,line width=0.65pt] (0,0)--(9pt,0);
  Sentinel-2 B7 (39.4)};
\end{tikzpicture}
\endgroup
\caption{\textbf{PRUE+ augmentation sweep.} Object F1 uses the released PRUE pixel-instance recipe on ten dense-label held-out countries, with an EfficientNet-B3 backbone, watershed, and D4 test-time augmentation. The first four bars use the CC-BY country subset and hold out several training countries; the All Countries bar uses the same recipe with all available training countries. Vertical reference lines denote Sentinel-2 baseline performance under the same recipe.}
    \label{fig:aug_ablation}
\end{figure}

\begin{figure*}[t!]
    \centering
    \begin{minipage}[t]{0.30\linewidth}
        \vspace{0pt}
        \centering
        \includegraphics[width=\linewidth]{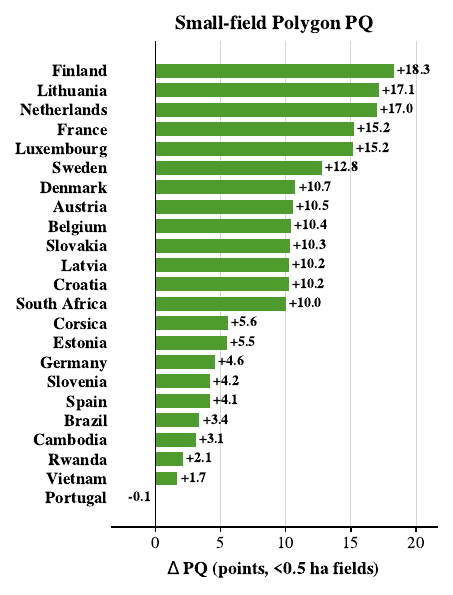}
    \end{minipage}\hspace{-0.02\linewidth}
    \begin{minipage}[t]{0.30\linewidth}
        \vspace{0pt}
        \centering
        \includegraphics[width=\linewidth]{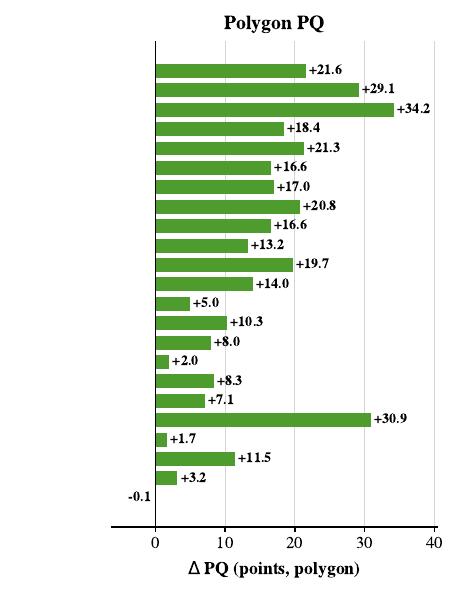}
    \end{minipage}\hspace{-0.02\linewidth}
    \begin{minipage}[t]{0.30\linewidth}
        \vspace{0pt}
        \centering
        \includegraphics[width=\linewidth]{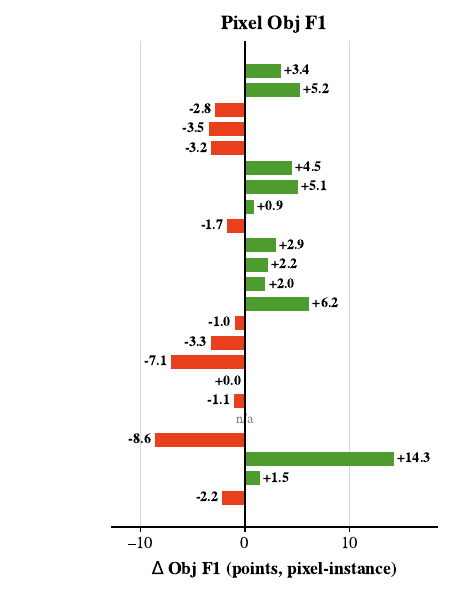}
    \end{minipage}
    \caption{\textbf{Regional cross-sensor comparison by metric.} Deltas are PlanetScope (B3) minus Sentinel-2 (B7). Left: polygon PQ for fields smaller than $0.5$\,ha. Center: polygon PQ for all fields. Right: FTW pixel-instance object F1 from released PRUE checkpoints~\cite{muhawenayo2026prue}. Kenya and India are excluded, Corsica is scored as its own region, and the PlanetScope pixel-instance F1 result is unavailable for Brazil. Polygon PQ favors PlanetScope in 22 of 23 regions and effectively ties in Portugal; pixel F1 favors PlanetScope in 11 of 22 comparable regions. Polygon-level improvements do not consistently translate into pixel-level gains.}
    \label{fig:per_country_bars}
\end{figure*}

\subsection{Experimental setup}
We follow the PRUE procedure~\cite{muhawenayo2026prue} and the official FTW per-region train/validation/test splits. Models are trained on the training split, selected on the validation split, and evaluated only on held-out test patches. Sentinel-2 and PlanetScope experiments use the same splits and evaluation code. Their predictions are vectorized from their native grids and scored against the same original FTW polygons.

We report averages over ten dense-label held-out countries: Belgium, Cambodia, Croatia, Germany, Latvia, Lithuania, Portugal, Slovenia, South Africa, and Sweden. Kenya is excluded because it is the only presence-only FTW region, where only a subset of fields is annotated, leaving the background class untrusted for supervised metrics. Portugal is retained because its labels are trusted, though its terraced micro-fields remain hard to resolve when parcels approach PlanetScope's ground sample distance. We return to Portugal in \Cref{sec:analysis}.

\paragraph{Training and selection.}
For both sensors, each model is a U-Net~\cite{ronneberger2015unet} with an EfficientNet encoder~\cite{tan2019efficientnet}, eight input channels from the two seasonal windows, and three output classes: background, field interior, and field boundary. We train with {\small\texttt{logcoshdice}} loss~\cite{jadon2020losses,milletari2016vnet}, class weights $[0.05,0.20,0.75]$, and AdamW~\cite{loshchilov2019adamw} with a learning rate of $10^{-3}$. Training runs for 100 epochs using bf16 mixed precision~\cite{micikevicius2018mixedprec}, with batch sizes of 32 for EfficientNet-B3 and 8 for EfficientNet-B7.

We select checkpoints using the loss on the FTW validation patches. PRUE+ uses the same architecture and optimizer as PRUE, adding the image and geometry augmentations listed in \Cref{app:heldout_sweep}, watershed (WS), and D4 test-time augmentation (TTA). The same recipe is applied to all reported Sentinel-2 and PlanetScope baselines. \Cref{fig:aug_ablation} and \Cref{tab:heldout} (appendix) report the corresponding sweeps.

\subsection{Controlled sensor comparison}
For a U-Net with an EfficientNet-B3 encoder, replacing 10\,m Sentinel-2 with 3\,m PlanetScope raises PQ from 21.0 to 35.5 and RQ from 28.9 to 46.2 (\Cref{tab:polygon_metrics}). The gain persists when parcel-level misses and hallucinations are counted, not only when the shapes of successful matches are scored. Mean boundary error on matched polygons falls from 18.6\,m to 7.4\,m. The PlanetScope B3 model also scores at or above the larger Sentinel-2 B7 baseline on the main polygon metrics. \Cref{fig:qualitative} shows held-out examples in which the Sentinel-2 baseline merges neighboring parcels that the PlanetScope model separates.

\subsection{Recipe analysis}

\paragraph{Augmentation.}
Augmentation gives the largest gain in the PlanetScope recipe (\Cref{fig:aug_ablation}). Combining the released PRUE preprocessing, seasonal-window swap and gamma jitter, and the PRUE+ geometry/noise recipe raises object F1 from 29.1 to 37.4; the geometry/noise recipe contributes the largest single increase. The sweep trains on the CC-BY subset and evaluates on countries absent from training, so it measures cross-region transfer. The final FTP-PRUE+ model uses the same recipe with all available training countries, reaching 46.2 polygon RQ in \Cref{tab:polygon_metrics} and 45.2 object F1 under the PRUE pixel-instance procedure in \Cref{fig:aug_ablation}. \Cref{app:heldout_sweep} lists the exact augmentation ranges.

\begin{figure*}[t!]
    \centering
    \includegraphics[width=0.85\linewidth]{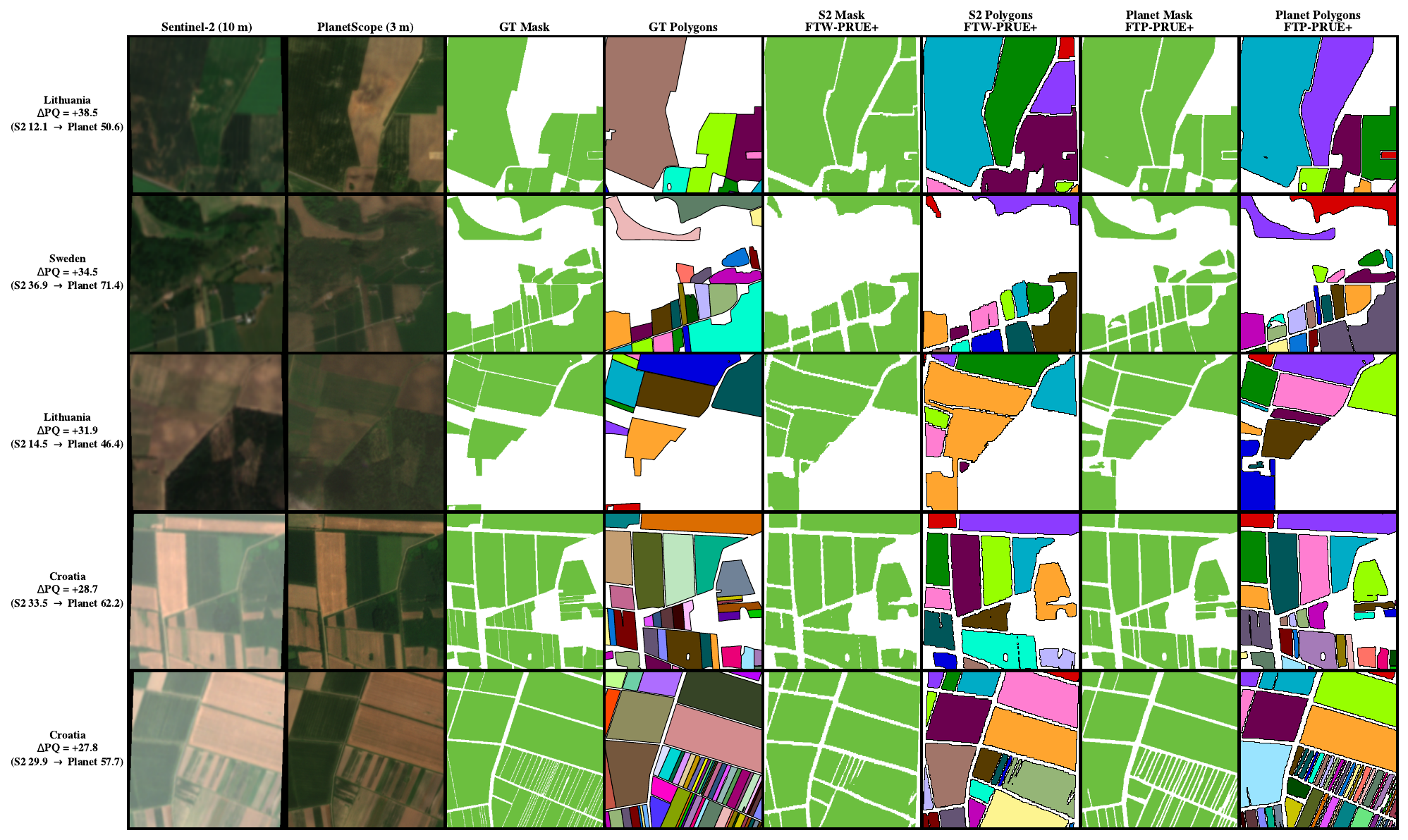}
    \caption{\textbf{Randomly sampled high-gain patches from 3\,m imagery.}
    Held-out examples sampled uniformly from the top 250 patches by FTW$\rightarrow$FTP polygon-PQ improvement. Columns show the 10\,m Sentinel-2 image, the corresponding 3\,m PlanetScope image, and mask/polygon pairs for the ground truth and each model. Per-patch PQ gain is shown at left. In these dense smallholder scenes, the FTW model tends to merge neighboring parcels or leave gaps, while the FTP model recovers more of the parcel structure.}
    \label{fig:improvement}
\end{figure*}

\paragraph{Inference refinements.}
Watershed separates touching field interiors before vectorization using $h$-maxima of the predicted boundary class's Euclidean distance transform as seeds. Watershed adds $+0.4$--$0.8$ object-F1 points across checkpoints. D4 test-time augmentation averages predictions over the eight flip-and-rotation symmetries~\cite{moshkov2020tta,wang2019tta}. It has little effect on the PRUE baseline but adds $+0.5$--$1.1$ points to PRUE+ checkpoints.

\paragraph{Backbone scaling.}
Increasing the PlanetScope backbone from EfficientNet-B3 to EfficientNet-B7 does not improve the dense held-out polygon metrics. The two are effectively tied on PQ, RQ, and boundary error (\Cref{tab:polygon_metrics}). The larger backbone adds compute without substantial gains.

\paragraph{Failed experiments.}
The PRUE+ geometry/noise recipe adds $+4.5$ object-F1 points over the lighter swap-order/gamma recipe, while the remaining specialized components have null or negative results (\Cref{tab:ablation_summary}, appendix). CutMix is neutral ($-0.1$), frame fields reduce object F1 by $1.7$ points, soft clDice fails to train, and the signed distance function (SDF) auxiliary head reduces object F1 by $3.1$ points. The SDF head helps without augmentation but hurts once PRUE+ is applied. This differs from DECODE~\cite{waldner2021decode}, which uses joint prediction of fields, boundaries, and distance to the nearest boundary. We retain strong image/geometry augmentation, watershed, and D4 test-time augmentation.

\section{Analysis}
\label{sec:analysis}

\subsection{Metric and grid controls}
\label{sec:polygon-first}

\paragraph{Metric choice.}
Across the 23 evaluated regions, polygon PQ favors PlanetScope in 22 and yields an effective tie in Portugal ($-0.1$ PQ). The released PRUE pixel-instance object-F1 comparison instead favors PlanetScope in only 11 of 22 comparable regions; Brazil lacks a PlanetScope pixel-instance result (\Cref{fig:per_country_bars}). The apparent benefit of PlanetScope thus depends on how field boundaries are evaluated. Polygon matching penalizes merged, fragmented, missed, and hallucinated parcels, whereas pixel overlap rewards area coverage even if the parcel objects are unusable (\Cref{fig:metric_example}).

\paragraph{Output grid.}
Native-grid evaluation also changes the raster grid, so we isolate this factor with two Sentinel-2 controls that provide a finer 512-pixel output grid without new image information: bilinear upsampling at evaluation only, and retraining on bilinearly upsampled Sentinel-2 imagery. All rows use the same prediction$\rightarrow$post-processing$\rightarrow$vectorization pipeline as the main polygon evaluation.

Retraining with upsampled Sentinel-2 raises PQ from 23.8 to 31.5, approaching PlanetScope's 36.0 (\Cref{tab:upsampled_s2_main}, appendix). These rows share the same scoring setup for the upsampling comparison, so their absolute values differ slightly from \Cref{tab:polygon_metrics}. The gain shows that output-grid resolution matters for parcel separation, but it does not recover boundary accuracy. The two finer-grid Sentinel-2 controls have 31.3\,m and 28.0\,m matched-boundary error, compared with 7.4\,m for PlanetScope. The PlanetScope result is therefore not explained by output-grid resolution alone.

\subsection{Regional and size diagnostics}

\paragraph{Regional variation.}
Across all evaluated regions, FTP U-Net EfficientNet-B3 reaches 38.5 macro-average PQ, compared with 24.2 for the larger FTW U-Net EfficientNet-B7 baseline, a $+14.3$ point margin (\Cref{fig:per_country_bars}). Both sensors are scored against the original FTW polygons at their native resolutions, but this average covers all 23 regions rather than the ten dense-label countries, so it is not directly comparable to \Cref{tab:polygon_metrics}.

Portugal is the only region without an FTP gain, with an effective tie at $-0.1$ PQ. Cambodia ($+1.7$ PQ) and Germany ($+2.0$ PQ) have the narrowest positive margins, although Germany's margin increases to $+4.6$ PQ on fields smaller than $0.5$\,ha. In Portugal and Germany, higher resolution alone cannot recover low-contrast cadastral boundaries that lack visible edges in the input imagery.

\begin{table}[t]
    \centering
    \caption{\textbf{Output-grid control by field size.}
    Upsampling improves Sentinel-2 PQ in every field-size group, but PlanetScope scores higher in all three. PQ (higher is better) is scored against the ground-truth FTW polygons. The Sentinel-2 rows use the same model at the original 10\,m resolution or with bilinear upsampling to a 512-pixel image size. PlanetScope uses 3\,m imagery.}
    \label{tab:resolution_ablation}
    \footnotesize
\setlength{\tabcolsep}{5pt}
\begin{tabular}{@{}lccc@{}}
\toprule
 & \multicolumn{3}{c}{PQ by field area} \\
\cmidrule(lr){2-4}
Condition & \makecell{Small\\($<0.5$\,ha)} & \makecell{Medium\\($0.5$--$2$\,ha)} & \makecell{Large\\($\ge2$\,ha)} \\
\midrule
Sentinel-2 (10\,m) & 5.8 & 25.3 & 33.8 \\
Sentinel-2 (upsampled) & 11.8 & 33.7 & 36.1 \\
\textbf{PlanetScope (3\,m)} & \textbf{15.7} & \textbf{39.2} & \textbf{52.0} \\
\bottomrule
\end{tabular}

\end{table}

\paragraph{Field size.}
We assign each ground-truth polygon to a size bin using its true vector area, then score each sensor against the original FTW polygons. FTP leads in every bin (\Cref{tab:resolution_ablation}). The largest relative gain occurs below $0.5$\,ha, where PlanetScope more than doubles PQ over native Sentinel-2 ($15.7$ vs.\ $5.8$). Upsampling Sentinel-2 raises small-field PQ to 11.8 but still trails PlanetScope, and the same ordering holds for medium and large fields. Together with the rasterization ceiling in \Cref{tab:representation_ceiling}, the size breakdown shows that a finer grid explains only part of the PlanetScope advantage. \Cref{fig:improvement} shows held-out examples sampled uniformly from the 250 patches with the largest FTW$\rightarrow$FTP polygon-PQ gains.

\section{Discussion and limitations}
\label{sec:limitations}

For smallholder mapping, a field merged with its neighbor is a missing parcel even if most crop pixels are classified correctly. PlanetScope's largest gains occur on these small fields, where polygon metrics capture improvements in parcel separation that pixel IoU understates. Optimizing pixel IoU alone can therefore miss progress in separating individual fields. FTP provides a way to test whether changes to the imagery or model produce more accurate parcel maps.

\paragraph{Scope of the findings.}
The rasterization test in \Cref{tab:representation_ceiling} counts parcels that remain separate in the labels before any model is trained. In the cross-sensor comparison, we hold the benchmark recipe in \Cref{sec:benchmark} fixed while changing the imagery source and raster grid. However, differences between sensors extend beyond ground sample distance to include radiometry, spectral content, and acquisition time. Upsampling tests whether more output pixels help separate fields without adding new observations. We compare U-Nets trained with the same recipe and the released Delineate Anything models. We build on PRUE's architecture comparison, where a U-Net with an EfficientNet encoder achieved higher object F1 than the evaluated transformer-based models~\cite{muhawenayo2026prue}, rather than repeating that architecture sweep.

\paragraph{Coverage and selection bias.}
FTP drops 7{,}800 of 140{,}968 (5.5\%) patch-window targets from the FTW dataset because no acceptable cloud-free PlanetScope scene is available. This primarily affects Portugal, Finland, and the Netherlands.

\paragraph{Cost and accessibility.}
Sentinel-2 is free and open, whereas PlanetScope imagery normally requires a purchased or program-based license. FTP releases the scene identifiers, timestamps, and download code rather than redistributing the paired imagery. Reproducing the benchmark therefore requires appropriate Planet access and remains subject to imagery costs and licensing constraints. Intended use, linkage risk, provenance, and geographic bias are detailed in \Cref{app:impact}.

\section{Conclusion}

Fields of the Planet is a high-resolution benchmark for testing how sensor choice, raster grids, and evaluation design affect field-boundary mapping. With one training recipe for both sensors, 3\,m PlanetScope raises PQ from 21.0 to 35.5 over 10\,m Sentinel-2 and reduces matched-boundary error from 18.6\,m to 7.4\,m, with the largest gains on fields under $0.5$\,ha. Rasterization and upsampling controls separate grid effects from image information, and polygon metrics measure parcel recovery rather than pixel overlap. These results support evaluating smallholder field maps by how well they separate parcels and locate their boundaries.

\section*{Acknowledgments}
This research used the TGI RAILs advanced compute and data resource which is supported by the National Science Foundation (award OAC-2232860) and the Taylor Geospatial.

{\small
\bibliographystyle{ieeenat_fullname}
\bibliography{refs}

\begin{thebibliography}{37}
\providecommand{\natexlab}[1]{#1}
\providecommand{\url}[1]{\texttt{#1}}
\expandafter\ifx\csname urlstyle\endcsname\relax
  \providecommand{\doi}[1]{doi: #1}\else
  \providecommand{\doi}{doi: \begingroup \urlstyle{rm}\Url}\fi

\bibitem[Astruc et~al.(2024)Astruc, Gonthier, Mallet, and Landrieu]{pastishd}
Guillaume Astruc, Nicolas Gonthier, Cl{\'e}ment Mallet, and Lo{\"i}c Landrieu.
\newblock {OmniSAT}: Self-supervised modality fusion for earth observation.
\newblock In \emph{European Conference on Computer Vision (ECCV)}, pages
  409--427, 2024.

\bibitem[Becker-Reshef et~al.(2020)Becker-Reshef, Justice, Barker, Humber,
  Rembold, Bonifacio, Zappacosta, Budde, Magadzire, Shitote, Pound,
  Constantino, Nakalembe, Mwangi, Sobue, Newby, Whitcraft, Jarvis, and
  Verdin]{beckerreshef2020geoglam}
Inbal Becker-Reshef, Christina Justice, Brian Barker, Michael Humber, Felix
  Rembold, Rogerio Bonifacio, Mario Zappacosta, Mike Budde, Tamuka Magadzire,
  Chris Shitote, Jonathan Pound, Alessandro Constantino, Catherine Nakalembe,
  Kenneth Mwangi, Shinichi Sobue, Terence Newby, Alyssa Whitcraft, Ian Jarvis,
  and James Verdin.
\newblock Strengthening agricultural decisions in countries at risk of food
  insecurity: The {GEOGLAM} crop monitor for early warning.
\newblock \emph{Remote Sensing of Environment}, 237:\penalty0 111553, 2020.

\bibitem[Bischke et~al.(2019)Bischke, Helber, Folz, Borth, and
  Dengel]{bischke2019buildings}
Benjamin Bischke, Patrick Helber, Joachim Folz, Damian Borth, and Andreas
  Dengel.
\newblock Multi-task learning for segmentation of building footprints with deep
  neural networks.
\newblock In \emph{IEEE International Conference on Image Processing (ICIP)},
  pages 1480--1484, 2019.

\bibitem[Cheng et~al.(2021)Cheng, Girshick, Doll{\'a}r, Berg, and
  Kirillov]{cheng2021biou}
Bowen Cheng, Ross Girshick, Piotr Doll{\'a}r, Alexander~C. Berg, and Alexander
  Kirillov.
\newblock Boundary {IoU}: Improving object-centric image segmentation
  evaluation.
\newblock In \emph{IEEE/CVF Conference on Computer Vision and Pattern
  Recognition (CVPR)}, pages 15334--15342, 2021.

\bibitem[Defourny et~al.(2019)Defourny, Bontemps, Bellemans, Cara, Dedieu,
  Guzzonato, Hagolle, Inglada, Nicola, Rabaute, Savinaud, Udroiu, Valero,
  B\'egu\'e, Dejoux, El~Harti, Ezzahar, Kussul, Labbassi, Lebourgeois, Miao,
  Newby, Nyamugama, Salh, Shelestov, Simonneaux, Traore, Traore, and
  Koetz]{defourny2019sen2agri}
Pierre Defourny, Sophie Bontemps, Nicolas Bellemans, Cosmin Cara, G\'erard
  Dedieu, Eric Guzzonato, Olivier Hagolle, Jordi Inglada, Laurentiu Nicola,
  Thierry Rabaute, Mickael Savinaud, Cosmin Udroiu, Silvia Valero, Agn\`es
  B\'egu\'e, Jean-Fran\c{c}ois Dejoux, Abderrazak El~Harti, Jamal Ezzahar,
  Nataliia Kussul, Kamal Labbassi, Valentine Lebourgeois, Zhang Miao, Terrence
  Newby, Adolph Nyamugama, Norakhan Salh, Andrii Shelestov, Vincent Simonneaux,
  Pierre~Sibiry Traore, Souleymane~S. Traore, and Benjamin Koetz.
\newblock Near real-time agriculture monitoring at national scale at parcel
  resolution: Performance assessment of the {Sen2-Agri} automated system in
  various cropping systems around the world.
\newblock \emph{Remote Sensing of Environment}, 221:\penalty0 551--568, 2019.

\bibitem[Drusch et~al.(2012)Drusch, Del~Bello, Carlier, Colin, Fernandez,
  Gascon, Hoersch, Isola, Laberinti, Martimort, Meygret, Spoto, Sy, Marchese,
  and Bargellini]{drusch2012sentinel2}
M. Drusch, U. Del~Bello, S. Carlier, O. Colin, V. Fernandez, F. Gascon, B.
  Hoersch, C. Isola, P. Laberinti, P. Martimort, A. Meygret, F. Spoto, O. Sy,
  F. Marchese, and P. Bargellini.
\newblock {Sentinel-2}: {ESA}'s optical high-resolution mission for {GMES}
  operational services.
\newblock \emph{Remote Sensing of Environment}, 120:\penalty0 25--36, 2012.

\bibitem[Estes et~al.(2024)Estes, Wussah, Asipunu, Gathigi,
  Kova{\v{c}}i{\v{c}}, Muhando, Yeboah, Addai, Akakpo, Allotey, Amkoya,
  Amponsem, Donkoh, Ha, Heltzel, Juma, Mdawida, Miroyo, Mucha, Mugami, Mwawaza,
  Nyarko, Oduor, Ohemeng, Segbefia, Tumbula, Wambua, Xeflide, Ye, and
  Yeboah]{estes2024africa}
Lyndon~D. Estes, Amos Wussah, Mary Asipunu, Michelle Gathigi, Primo{\v{z}}
  Kova{\v{c}}i{\v{c}}, Justus Muhando, B.~V. Yeboah, F.~K. Addai, E.~S. Akakpo,
  M.~K. Allotey, P. Amkoya, E. Amponsem, K.~D. Donkoh, N. Ha, E. Heltzel, C.
  Juma, R. Mdawida, A. Miroyo, J. Mucha, J. Mugami, F. Mwawaza, D.~A. Nyarko,
  P. Oduor, K.~N. Ohemeng, S.~I.~D. Segbefia, T. Tumbula, F. Wambua, G.~H.
  Xeflide, S. Ye, and F. Yeboah.
\newblock A region-wide, multi-year set of crop field boundary labels for
  africa.
\newblock arXiv preprint arXiv:2412.18483, 2024.

\bibitem[Garc\'ia-Pedrero et~al.(2017)Garc\'ia-Pedrero, Gonzalo-Mart\'in, and
  Lillo-Saavedra]{garciapedrero2017boundary}
Angel Garc\'ia-Pedrero, Consuelo Gonzalo-Mart\'in, and Mario Lillo-Saavedra.
\newblock A machine learning approach for agricultural parcel delineation
  through agglomerative segmentation.
\newblock \emph{International Journal of Remote Sensing}, 38\penalty0
  (7):\penalty0 1809--1819, 2017.

\bibitem[Girard et~al.(2021)Girard, Smirnov, Solomon, and
  Tarabalka]{girard2021ffl}
Nicolas Girard, Dmitriy Smirnov, Justin Solomon, and Yuliya Tarabalka.
\newblock Polygonal building extraction by frame field learning.
\newblock In \emph{IEEE/CVF Conference on Computer Vision and Pattern
  Recognition (CVPR)}, pages 5891--5900, 2021.

\bibitem[Hayder et~al.(2017)Hayder, He, and Salzmann]{hayder2017bais}
Zeeshan Hayder, Xuming He, and Mathieu Salzmann.
\newblock Boundary-aware instance segmentation.
\newblock In \emph{IEEE Conference on Computer Vision and Pattern Recognition
  (CVPR)}, pages 587--595, 2017.

\bibitem[Jadon(2020)]{jadon2020losses}
Shruti Jadon.
\newblock A survey of loss functions for semantic segmentation.
\newblock In \emph{IEEE Conference on Computational Intelligence in
  Bioinformatics and Computational Biology (CIBCB)}, pages 1--7, 2020.

\bibitem[Kerner et~al.(2025)Kerner, Chaudhari, Ghosh, Robinson, Ahmad, Choi,
  Jacobs, Holmes, Mohr, Dodhia, Lavista~Ferres, and Marcus]{kerner2024ftw}
Hannah Kerner, Snehal Chaudhari, Aninda Ghosh, Caleb Robinson, Adeel Ahmad,
  Eddie Choi, Nathan Jacobs, Chris Holmes, Matthias Mohr, Rahul Dodhia, Juan~M.
  Lavista~Ferres, and Jennifer Marcus.
\newblock Fields of the world: A machine learning benchmark dataset for global
  agricultural field boundary segmentation.
\newblock In \emph{Proceedings of the AAAI Conference on Artificial
  Intelligence}, pages 28151--28159, 2025.

\bibitem[{Kerner Lab}(2024)]{ftw-source-coop}
{Kerner Lab}.
\newblock Fields of the world: Field boundary {GeoParquet} releases.
\newblock \url{https://source.coop/kerner-lab}, 2024.

\bibitem[Khallaghi et~al.(2025)Khallaghi, Abedi, Abou~Ali, Alemohammad,
  Dziedzorm~Asipunu, Alatise, Ha, Luo, Mai, Song, Wussah, Xiong, Yao, Zhang,
  and Estes]{khallaghi2025generalization}
Sam Khallaghi, Rahebeh Abedi, Hanan Abou~Ali, Hamed Alemohammad, Mary
  Dziedzorm~Asipunu, Ismail Alatise, Nguyen Ha, Boka Luo, Cat Mai, Lei Song,
  Amos~Olertey Wussah, Sitian Xiong, Yao-Ting Yao, Qi Zhang, and Lyndon~D.
  Estes.
\newblock Generalization enhancement strategies to enable cross-year cropland
  mapping with convolutional neural networks trained using historical samples.
\newblock \emph{Remote Sensing}, 17\penalty0 (3):\penalty0 474, 2025.

\bibitem[Kirillov et~al.(2019)Kirillov, He, Girshick, Rother, and
  Doll{\'a}r]{kirillov2019panoptic}
Alexander Kirillov, Kaiming He, Ross Girshick, Carsten Rother, and Piotr
  Doll{\'a}r.
\newblock Panoptic segmentation.
\newblock In \emph{IEEE/CVF Conference on Computer Vision and Pattern
  Recognition (CVPR)}, 2019.

\bibitem[Lavreniuk et~al.(2025)Lavreniuk, Kussul, Shelestov, Yailymov, Salii,
  Kuzin, and Szantoi]{lavreniuk2025delineate}
Mykola Lavreniuk, Nataliia Kussul, Andrii Shelestov, Bohdan Yailymov, Yevhenii
  Salii, Volodymyr Kuzin, and Zoltan Szantoi.
\newblock Delineate anything: Resolution-agnostic field boundary delineation on
  satellite imagery.
\newblock arXiv preprint arXiv:2504.02534, 2025.

\bibitem[Lavreniuk et~al.(2026)Lavreniuk, Kussul, Shelestov, Salii, Kuzin,
  Wang, and Szantoi]{lavreniuk2026delineatev2}
Mykola Lavreniuk, Nataliia Kussul, Andrii Shelestov, Yevhenii Salii, Volodymyr
  Kuzin, Charlotte Julia Li-Xing Wang, and Zoltan Szantoi.
\newblock Delineate anything v2: A global foundation model for field
  delineation.
\newblock arXiv preprint arXiv:2607.19069, 2026.

\bibitem[Lin et~al.(2014)Lin, Maire, Belongie, Hays, Perona, Ramanan,
  Doll{\'a}r, and Zitnick]{lin2014coco}
Tsung-Yi Lin, Michael Maire, Serge Belongie, James Hays, Pietro Perona, Deva
  Ramanan, Piotr Doll{\'a}r, and C.~Lawrence Zitnick.
\newblock {Microsoft COCO}: Common objects in context.
\newblock In \emph{European Conference on Computer Vision (ECCV)}, pages
  740--755, 2014.

\bibitem[Loshchilov and Hutter(2019)]{loshchilov2019adamw}
Ilya Loshchilov and Frank Hutter.
\newblock Decoupled weight decay regularization.
\newblock In \emph{International Conference on Learning Representations
  (ICLR)}, 2019.

\bibitem[Lowder et~al.(2016)Lowder, Skoet, and Raney]{lowder2016farms}
Sarah~K. Lowder, Jakob Skoet, and Terri Raney.
\newblock The number, size, and distribution of farms, smallholder farms, and
  family farms worldwide.
\newblock \emph{World Development}, 87:\penalty0 16--29, 2016.

\bibitem[Lowder et~al.(2021)Lowder, S{\'a}nchez, and Bertini]{lowder2021farms}
Sarah~K. Lowder, Marco~V. S{\'a}nchez, and Raffaele Bertini.
\newblock Which farms feed the world and has farmland become more concentrated?
\newblock \emph{World Development}, 142:\penalty0 105455, 2021.

\bibitem[Micikevicius et~al.(2018)Micikevicius, Narang, Alben, Diamos, Elsen,
  Garc{\'i}a, Ginsburg, Houston, Kuchaiev, Venkatesh, and
  Wu]{micikevicius2018mixedprec}
Paulius Micikevicius, Sharan Narang, Jonah Alben, Gregory Diamos, Erich Elsen,
  David Garc{\'i}a, Boris Ginsburg, Michael Houston, Oleksii Kuchaiev, Ganesh
  Venkatesh, and Hao Wu.
\newblock Mixed precision training.
\newblock In \emph{International Conference on Learning Representations
  (ICLR)}, 2018.

\bibitem[Milletari et~al.(2016)Milletari, Navab, and Ahmadi]{milletari2016vnet}
Fausto Milletari, Nassir Navab, and Seyed-Ahmad Ahmadi.
\newblock {V-Net}: Fully convolutional neural networks for volumetric medical
  image segmentation.
\newblock In \emph{International Conference on 3D Vision (3DV)}, pages
  565--571, 2016.

\bibitem[Moshkov et~al.(2020)Moshkov, Math{\'e}, Kert{\'e}sz-Farkas, Hollandi,
  and Horv{\'a}th]{moshkov2020tta}
Nikita Moshkov, Botond Math{\'e}, Attila Kert{\'e}sz-Farkas, R{\'e}ka Hollandi,
  and P{\'e}ter Horv{\'a}th.
\newblock Test-time augmentation for deep learning-based cell segmentation on
  microscopy images.
\newblock \emph{Scientific Reports}, 10:\penalty0 5068, 2020.

\bibitem[Muhawenayo et~al.(2026)Muhawenayo, Robinson, Khanal, Fang, Corley,
  Wollam, Gao, Strnad, Avery, Estes, T{\'a}rano, Jacobs, and
  Kerner]{muhawenayo2026prue}
Gedeon Muhawenayo, Caleb Robinson, Subash Khanal, Zhanpei Fang, Isaac Corley,
  Alexander Wollam, Tianyi Gao, Leonard Strnad, Ryan Avery, Lyndon Estes,
  Ana~M. T{\'a}rano, Nathan Jacobs, and Hannah Kerner.
\newblock {PRUE}: A practical recipe for field boundary segmentation at scale.
\newblock In \emph{Proceedings of the IEEE/CVF Conference on Computer Vision
  and Pattern Recognition (CVPR)}, pages 6484--6495, 2026.

\bibitem[Persello et~al.(2019)Persello, Tolpekin, Bergado, and
  de~By]{persello2019delineation}
Claudio Persello, Valentyn~A. Tolpekin, John~R. Bergado, and Rolf~A. de By.
\newblock Delineation of agricultural fields in smallholder farms from
  satellite images using fully convolutional networks and combinatorial
  grouping.
\newblock \emph{Remote Sensing of Environment}, 231:\penalty0 111253, 2019.

\bibitem[Persello et~al.(2023)Persello, Grift, Fan, Paris, H{\"a}nsch, Koeva,
  and Nelson]{persello2023ai4smallfarms}
Claudio Persello, Jeroen Grift, Xinyan Fan, Claudia Paris, Ronny H{\"a}nsch,
  Mila Koeva, and Andrew Nelson.
\newblock {AI4SmallFarms}: A dataset for crop field delineation in southeast
  asian smallholder farms.
\newblock \emph{IEEE Geoscience and Remote Sensing Letters}, 20:\penalty0 1--5,
  2023.

\bibitem[{Planet Team}(2017)]{planet2017}
{Planet Team}.
\newblock Planet application program interface: In space for life on earth.
\newblock \url{https://api.planet.com}, 2017.

\bibitem[Riba et~al.(2020)Riba, Mishkin, Ponsa, Rublee, and
  Bradski]{riba2020kornia}
Edgar Riba, Dmytro Mishkin, Daniel Ponsa, Ethan Rublee, and Gary Bradski.
\newblock {Kornia}: an open source differentiable computer vision library for
  {PyTorch}.
\newblock In \emph{IEEE Winter Conference on Applications of Computer Vision
  (WACV)}, pages 3663--3672, 2020.

\bibitem[Ronneberger et~al.(2015)Ronneberger, Fischer, and
  Brox]{ronneberger2015unet}
Olaf Ronneberger, Philipp Fischer, and Thomas Brox.
\newblock {U-Net}: Convolutional networks for biomedical image segmentation.
\newblock In \emph{International Conference on Medical Image Computing and
  Computer-Assisted Intervention (MICCAI)}, pages 234--241, 2015.

\bibitem[Sainte Fare~Garnot and Landrieu(2021)]{garnot2021pastis}
Vivien Sainte Fare~Garnot and Lo{\"i}c Landrieu.
\newblock Panoptic segmentation of satellite image time series with
  convolutional temporal attention networks.
\newblock In \emph{IEEE/CVF International Conference on Computer Vision
  (ICCV)}, pages 4872--4881, 2021.

\bibitem[Shit et~al.(2021)Shit, Paetzold, Sekuboyina, Ezhov, Unger, Zhylka,
  Pluim, Bauer, and Menze]{shit2021cldice}
Suprosanna Shit, Johannes~C. Paetzold, Anjany Sekuboyina, Ivan Ezhov, Alexander
  Unger, Andrey Zhylka, Josien P.~W. Pluim, Ulrich Bauer, and Bjoern~H. Menze.
\newblock {clDice} -- a novel topology-preserving loss function for tubular
  structure segmentation.
\newblock In \emph{IEEE/CVF Conference on Computer Vision and Pattern
  Recognition (CVPR)}, pages 16560--16569, 2021.

\bibitem[Tan and Le(2019)]{tan2019efficientnet}
Mingxing Tan and Quoc~V. Le.
\newblock {EfficientNet}: Rethinking model scaling for convolutional neural
  networks.
\newblock In \emph{International Conference on Machine Learning (ICML)}, pages
  6105--6114, 2019.

\bibitem[Waldner et~al.(2021)Waldner, Diakogiannis, Batchelor, Ciccotosto-Camp,
  Cooper-Williams, Herrmann, Mata, and Toovey]{waldner2021decode}
Fran{\c{c}}ois Waldner, Foivos~I. Diakogiannis, Kathryn Batchelor, Michael
  Ciccotosto-Camp, Elizabeth Cooper-Williams, Chris Herrmann, Gonzalo Mata, and
  Andrew Toovey.
\newblock Detect, consolidate, delineate: Scalable mapping of field boundaries
  using satellite images.
\newblock \emph{Remote Sensing}, 13\penalty0 (11):\penalty0 2197, 2021.

\bibitem[Wang et~al.(2019)Wang, Li, Aertsen, Deprest, Ourselin, and
  Vercauteren]{wang2019tta}
Guotai Wang, Wenqi Li, Michael Aertsen, Jan Deprest, S{\'e}bastien Ourselin,
  and Tom Vercauteren.
\newblock Aleatoric uncertainty estimation with test-time augmentation for
  medical image segmentation with convolutional neural networks.
\newblock \emph{Neurocomputing}, 338:\penalty0 34--45, 2019.

\bibitem[Wang et~al.(2022)Wang, Waldner, and Lobell]{wang2022unlocking}
Sherrie Wang, Fran{\c{c}}ois Waldner, and David~B. Lobell.
\newblock Unlocking large-scale crop field delineation in smallholder farming
  systems with transfer learning and weak supervision.
\newblock \emph{Remote Sensing}, 14\penalty0 (22):\penalty0 5738, 2022.

\bibitem[Yun et~al.(2019)Yun, Han, Oh, Chun, Choe, and Yoo]{yun2019cutmix}
Sangdoo Yun, Dongyoon Han, Seong~Joon Oh, Sanghyuk Chun, Junsuk Choe, and
  Youngjoon Yoo.
\newblock {CutMix}: Regularization strategy to train strong classifiers with
  localizable features.
\newblock In \emph{IEEE/CVF International Conference on Computer Vision
  (ICCV)}, 2019.

\end{thebibliography}
}

\clearpage
\appendix
\crefalias{section}{appendix}
\setcounter{table}{0}
\setcounter{figure}{0}
\setcounter{equation}{0}
\renewcommand{\thetable}{S\arabic{table}}
\renewcommand{\thefigure}{S\arabic{figure}}
\renewcommand{\theequation}{S\arabic{equation}}
\renewcommand{\theHtable}{appendix.\arabic{table}}
\renewcommand{\theHfigure}{appendix.\arabic{figure}}
\renewcommand{\theHequation}{appendix.\arabic{equation}}
\section{Polygon-metric definitions}
\label{app:metrics}

All object-level metrics use the field-interior class. Each predicted mask is vectorized and compared with the original FTW ground-truth polygons. Matching and metric counts are pooled within a country before countries receive equal weight in the reported macro-average. Pixel IoU is computed on raster masks before vectorization. The equations express PQ, SQ, RQ, F1, and pixel IoU on a $[0,1]$ scale. Tables report these scores as percentages and polygon-count error as a ratio.

\paragraph{Vectorization and matching.}
For each patch, the post-processed field-interior prediction is converted into polygons $\mathcal{P}=\{p_1,\ldots,p_M\}$ and compared with the ground-truth polygons $\mathcal{G}=\{g_1,\ldots,g_N\}$. All intersections and unions use polygon areas from the vector geometries. At IoU threshold $\tau$, a predicted polygon and a ground-truth polygon are matched when
\begin{equation}
\mathrm{IoU}(g,p)=
\frac{\mathrm{area}(g\cap p)}
{\mathrm{area}(g\cup p)}
>\tau .
\end{equation}
For $\tau\ge0.5$, disjoint polygon interiors make the match set $\mathcal{M}_\tau$ a partial one-to-one matching. The released evaluator uses deterministic greedy matching: ground-truth polygons are visited in stored order, predictions in vectorization order, and each ground-truth polygon takes the first unmatched prediction with $\mathrm{IoU}>\tau$. Matching is recomputed independently at each threshold. We then define $\mathrm{TP}_\tau=|\mathcal{M}_\tau|$, $\mathrm{FP}_\tau=M-\mathrm{TP}_\tau$, and $\mathrm{FN}_\tau=N-\mathrm{TP}_\tau$.

\paragraph{Object F1 and panoptic quality.}
At threshold $\tau$, object F1 is $2\mathrm{TP}_\tau/(2\mathrm{TP}_\tau+\mathrm{FP}_\tau+\mathrm{FN}_\tau)$; the evaluator returns zero when the denominator is zero. We report F1 at $\mathrm{IoU}=0.5$ and its mean over the COCO threshold grid $\mathcal{T}=\{0.5,0.55,\ldots,0.95\}$~\cite{lin2014coco}:
\begin{equation}
\mathrm{F1}_{[.5:.95]}=
\frac{1}{|\mathcal{T}|}
\sum_{\tau\in\mathcal{T}}\mathrm{F1}_{\tau}.
\end{equation}
Panoptic quality follows Kirillov et al.~\cite{kirillov2019panoptic}. With matches fixed at $\tau=0.5$, recognition quality is $\mathrm{RQ}=\mathrm{F1}_{0.5}$, segmentation quality is the mean IoU of matched polygons,
\begin{equation}
\mathrm{SQ}=
\frac{1}{\mathrm{TP}_{0.5}}
\sum_{(g,p)\in\mathcal{M}_{0.5}}
\mathrm{IoU}(g,p),
\end{equation}
and $\mathrm{PQ}=\mathrm{SQ}\times\mathrm{RQ}$. If $\mathrm{TP}_{0.5}=0$, we set $\mathrm{SQ}=0$ and $\mathrm{PQ}=0$, while matched-boundary error is undefined. RQ and F1 still penalize false positives and false negatives in this case.

\paragraph{Polygon-count error.}
Let $\bar{M}$ and $\bar{N}$ denote the mean numbers of predicted and ground-truth polygons per patch within a country. We report the normalized count discrepancy
\begin{equation}
|\Delta N|/N=
\frac{|\bar{M}-\bar{N}|}{\bar{N}},
\end{equation}
computed per country and then macro-averaged. This metric compares polygon counts without IoU matching and treats overprediction and underprediction equally.

\paragraph{Boundary error.}
For each matched pair $(g,p)\in\mathcal{M}_{0.5}$, we compute a symmetric boundary Chamfer distance between one-pixel-wide ground-truth and predicted boundary sets on the evaluated raster grid. Euclidean distance transforms give the nearest-boundary distance in each direction; we average the two directions and convert pixels to meters using the evaluated grid's ground sample distance, 3\,m for PlanetScope and 10\,m for Sentinel-2. A matched pair with an empty boundary set has undefined Chamfer distance and is excluded. Within each country, we report the mean and 95th percentile over defined matched pairs. Boundary error measures localization for matched fields, while RQ and object F1 penalize missed and hallucinated fields.

\paragraph{Pixel IoU.}
We report the pixel-level Jaccard index for the field-interior class,
\begin{equation}
\mathrm{IoU}_{\mathrm{px}} =
\frac{|\hat{Y}_{\mathrm{field}}\cap Y_{\mathrm{field}}|}
{|\hat{Y}_{\mathrm{field}}\cup Y_{\mathrm{field}}|},
\end{equation}
computed on raster masks before vectorization. We compute pixel IoU from the total intersection and union counts in each country, then average the country scores. Unlike the polygon metrics, pixel IoU has no notion of individual parcels and is not comparable across resolutions.

\paragraph{Aggregation.}

\begin{table}[ht!]
\centering
\small
\caption{\textbf{Dataset scope.}
FTP contains 66{,}584 complete two-window patch pairs across FTW's 24 countries and 25 labeled regions. \emph{Success rate} is the fraction of all 140{,}968 FTW patch-window targets paired with a PlanetScope image (133{,}168; 94.5\%). UDM2 statistics are available for 129{,}490 of these images.}
\label{tab:scope}
\begin{tabular}{lr}
\toprule
\textbf{Property} & \textbf{Value} \\
\midrule
\multicolumn{2}{l}{\cellcolor{black!7}\textbf{Geographic coverage}}\\
Countries / regions / continents & 24\,/\,25\,/\,4 \\
FTW patches                  & 70{,}484 \\
(Patch, window) targets      & 140{,}968 \\
\midrule
\multicolumn{2}{l}{\cellcolor{black!7}\textbf{Source imagery}}\\
Distinct source scenes & 6{,}113 \\
\midrule
\multicolumn{2}{l}{\cellcolor{black!7}\textbf{Constructed dataset}}\\
PlanetScope SR (4-band, uint16) & 133{,}168 \\
Per-window UDM2 QA statistics   & 129{,}490 \\
3-class labels (uint8, 1/patch) & 66{,}584 \\
\midrule
\multicolumn{2}{l}{\cellcolor{black!7}\textbf{Yield}}\\
Successfully matched targets    & 133{,}168 \\
\textbf{Success rate}           & \textbf{94.5\%} \\
Total dataset size              & 102\,GB \\
\bottomrule
\end{tabular}
\end{table}

We first pool $\mathrm{TP}/\mathrm{FP}/\mathrm{FN}$ counts and matched IoUs across patches within each country, then compute RQ, SQ, PQ, and $\mathrm{F1}_{[.5:.95]}$. Chamfer means and 95th percentiles are also computed within each country. \Cref{tab:polygon_metrics} then takes the arithmetic mean over the ten dense-label held-out countries, giving each country equal weight. If a country has no matched polygons, its SQ and PQ are zero, while its boundary mean and percentile are undefined and excluded from the corresponding macro-average. All recognition and count columns remain defined. The reported boundary p95 is the mean of country-level 95th percentiles.

\paragraph{Field-area breakdown.}
For \Cref{tab:resolution_ablation}, ground-truth polygons are assigned by UTM area $A$ to \emph{small} ($A<0.5$\,ha), \emph{medium} ($0.5\le A<2$\,ha), or \emph{large} ($A\ge2$\,ha). The 2\,ha threshold follows the standard smallholder ceiling~\cite{lowder2016farms}, while 0.5\,ha separates the dominant sub-hectare fields in FTW. Each unmatched prediction is a false positive in the bin determined by its predicted area. Counts and matched IoUs are summed within each country, PQ is computed per bin, and the ten dense-label held-out countries are macro-averaged. A bin with no matched pair has SQ and PQ equal to zero.

\section{Upsampled Sentinel-2 control}
\label{app:upsampled_s2}

We test the effect of a finer prediction grid by bilinearly resizing the 256-pixel Sentinel-2 input to 512 pixels before inference without retraining. A second condition uses bilinearly resized 512-pixel Sentinel-2 imagery during both training and evaluation.

\begin{table}[htb]
    \centering
    \small
    \setlength{\tabcolsep}{4pt}
    \caption{\textbf{Sentinel-2 output-grid control.} B3 models use watershed and D4 TTA on the ten dense-label held-out countries. PQ, SQ, and RQ are percentage points (higher is better); matched-boundary error is lower is better. Rows share the same scoring setup for the upsampling comparison, using rasterized labels, so absolute scores differ from the true-polygon native-grid results in \Cref{tab:polygon_metrics}. Boundary statistics exclude countries without a matched pair; native Sentinel-2 boundary error is averaged over nine countries because it is undefined for Portugal.}
    \label{tab:upsampled_s2_main}
    \begin{tabular}{lcccc}
    \toprule
    Condition & PQ$\uparrow$ & SQ$\uparrow$ & RQ$\uparrow$ & Bd.\ err (m)$\downarrow$ \\
    \midrule
    Native ($256$, $10$\,m)         & $23.8$ & $67.8$ & $30.8$ & $18.6$ \\
    Upsampled at eval ($512$)     & $27.4$ & $75.8$ & $34.9$ & $31.3$ \\
    Trained upsampled ($512$)     & $31.5$ & $77.5$ & $39.5$ & $28.0$ \\
    \midrule
    Planet ($512$, $3$\,m)   & $\mathbf{36.0}$ & $\mathbf{77.9}$ & $\mathbf{45.2}$ & $\mathbf{7.4}$ \\
    \bottomrule
    \end{tabular}
\end{table}

\section{Dataset scope and per-tile UDM2 quality}
\label{app:scope}
\label{app:udm2_country}
\Cref{tab:scope} summarizes geographic coverage, source scenes, and pairing yield.

\begin{table}[ht!]
\centering
\small
\caption{\textbf{Per-tile UDM2 coverage.}
Each entry is the percentage of the 129{,}490 patch-window tiles with valid UDM2 statistics for which the named class covers any, at least 50\%, or at least 90\% of the tile. These columns describe the distribution; the strict usable-tile flag instead requires clear$\,\ge\,$95\% and unusable$\,\le\,$5\%, which 89.7\% of valid-QA tiles satisfy.}
\label{tab:udm2}
\begin{tabular}{l S[table-format=2.1] S[table-format=2.1] S[table-format=2.1]}
\toprule
{Band} & {any (\textgreater0)} & {\bfseries $\ge 50\%$} & {$\ge 90\%$} \\
\midrule
\texttt{clear}      & 98.9 & \bfseries 95.9 & 91.1 \\
\texttt{cloud}      &  9.0 & \bfseries  1.0 &  0.2 \\
\texttt{shadow}     &  6.9 & \bfseries  0.2 &  0.0 \\
\texttt{light\_haze}&  5.1 & \bfseries  2.0 &  1.1 \\
\texttt{unusable}   & 11.3 & \bfseries  1.3 &  0.4 \\
\bottomrule
\end{tabular}
\par\vspace{2pt}\footnotesize \% of valid-QA (patch, window) tiles; each entry thresholds the class area as a fraction of that tile.
\end{table}

\paragraph{Regional QA.}
Usable-tile rates vary widely by region, and field-label coverage varies independently of image quality; India, for example, has high clear-sky fractions but only $\sim$1\% field pixels per patch. The release includes the full per-region index as a CSV so users can compare image quality, label density, and dataset retention.

\section{Broader impact and ethical considerations}
\label{app:impact}

\paragraph{Intended use.}
FTP is intended for research on agricultural field-boundary delineation and for evaluating methods used in aggregate crop-area and yield estimation, irrigation and water accounting, subsidy-program monitoring, and food-security analysis. Its main intended benefit is better evaluation in smallholder regions, where a 10\,m raster cannot represent many parcels (\Cref{tab:representation_ceiling}). FTP is a benchmark, not a validated system for identifying growers, adjudicating ownership, determining individual credit or subsidy eligibility, or enforcing decisions against individual farms.

\begin{table}[htb]
    \centering
    \scriptsize
    \caption{\textbf{Watershed and D4-TTA sensitivity.}
    Object F1 (\%) uses the PRUE pixel-instance evaluation. PRUE+ results are averaged over ten dense-label test countries. Published PRUE results ($^\ddag$) use a different test split and instance-extraction method~\cite{muhawenayo2026prue}. WS is watershed, and TTA averages predictions over eight D4 transformations.}
    \label{tab:heldout}
    \footnotesize
\setlength{\tabcolsep}{3pt}
\renewcommand{\arraystretch}{1.05}
\begin{tabular}{@{}l l ccc c c@{}}
\toprule
& & \multicolumn{4}{c}{Obj F1 (10-country dense held-out)} & \\
\cmidrule(lr){3-6}
Model & Backbone & \makecell{no-WS\\no-TTA} & \makecell{no-WS\\TTA} & \makecell{WS\\no-TTA} & \makecell{\textbf{WS}\\\textbf{+TTA}} & \makecell{Pix\\IoU} \\
\midrule
\multicolumn{7}{@{}l}{\textit{FTW-PRUE, released by \cite{muhawenayo2026prue}}} \\
FTW-PRUE & B3 & --- & --- & --- & 43.0$^\ddag$ & 74.0$^\ddag$ \\
FTW-PRUE & B5 & --- & --- & --- & 46.0$^\ddag$ & 75.0$^\ddag$ \\
FTW-PRUE & B7 & --- & --- & --- & 47.0$^\ddag$ & 76.0$^\ddag$ \\
\midrule
\multicolumn{7}{@{}l}{\textit{Ours --- FTP-PRUE+}} \\
FTP-PRUE+ & B3 & 43.5 & 43.9 & 44.1 & \textbf{45.2} & 68.8 \\
\midrule
\multicolumn{7}{@{}l}{\textit{FTW-PRUE baselines retrained with our \textbf{PRUE+} recipe}} \\
FTW-PRUE+ & B3 & 33.4 & 33.3 & 33.9 & \textbf{34.9} & 61.8 \\
FTW-PRUE+ & B7 & 39.9 & 39.5 & 40.4 & \textbf{40.7} & 63.6 \\
\bottomrule
\end{tabular}
\end{table}

\paragraph{Privacy and surveillance risks.}
Parcel maps joined with ownership or registry data could be used to monitor, score, or target individual farms, and smallholders are the most exposed to this. FTP contains no ownership, identity, or personal-information fields, and its polygons were already published through FTW; we collected no new parcel records. Parcel geometry can still be linked to external records, so publishing FTP does not remove that risk. The CC-BY-NC 4.0 license limits commercial reuse but does not prevent non-commercial misuse. FTP is intended for aggregate monitoring and benchmark research, not for identifying, ranking, or profiling individual farms.

\paragraph{Provenance and consent.}
All labels retain the per-region FTW licenses under which governments and labeling campaigns published them, ranging from CC-BY to CC-BY-NC~\cite{kerner2024ftw}; FTP redistributes the geometries unchanged. PlanetScope imagery is \copyright{} Planet Labs PBC and was obtained directly from the Planet archive under a research license for academic and nonprofit use. Use and redistribution of Planet imagery remain subject to the applicable Planet license terms.

\paragraph{Geographic and selection bias.}

\begin{table}[ht!]
    \centering
    \footnotesize
    \setlength{\tabcolsep}{3pt}
    \caption{\textbf{Recipe component sensitivity.}
    Changes in object F1 are averaged over ten dense-label test countries, using WS and TTA unless noted. The first three rows add augmentations in sequence on the CC-BY subset. Other rows compare separate experiments and should not be added together. A checkmark indicates a component kept in PRUE+, $\times$ indicates one excluded, and ``failed'' means training diverged.}
    \label{tab:ablation_summary}
    \begin{tabular}{@{}p{0.62\linewidth} @{\hspace{2pt}}c@{\hspace{2pt}} r@{}}
    \toprule
    \textbf{Component} & \textbf{Use} & \textbf{$\Delta$ ObjF1 (pts)} \\
    \midrule
    \multicolumn{3}{@{}l}{\cellcolor{black!7}\textbf{Augmentation \& training}} \\
    PRUE preprocess+resize aug.~\cite{muhawenayo2026prue} & \checkmark & {\color{earthDark}\bfseries +2.8} \\
    \texttt{swap\_order} + per-band $\gamma$ & \checkmark & {\color{earthDark}\bfseries +1.0} \\
    Geometry/noise recipe (PRUE+: affine, blur, noise, window dropout, rot, shear, jitter) & \checkmark & {\color{earthDark}\bfseries +4.5} \\
    Backbone B3 $\to$ B7 (no gain, larger) & $\times$ & {\color{black!50}\bfseries $\sim$0} \\
    Soft clDice~\cite{shit2021cldice} on boundary & $\times$ & {\color{earthSienna}\bfseries failed} \\
    SDF auxiliary head~\cite{bischke2019buildings,hayder2017bais} (full-data PRUE+) & $\times$ & {\color{earthSienna}\bfseries -3.1} \\
    Frame-field head~\cite{girard2021ffl} & $\times$ & {\color{earthSienna}\bfseries -1.7} \\
    CutMix~\cite{yun2019cutmix} (2\,px ignore) & $\times$ & {\color{earthSienna}\bfseries -0.1} \\
    \midrule
    \multicolumn{3}{@{}l}{\cellcolor{black!7}\textbf{Inference \& post-processing}} \\
    Marker-controlled watershed & \checkmark & {\color{earthDark}\bfseries +0.5} \\
    D4 8-way TTA (with PRUE+) & \checkmark & {\color{earthDark}\bfseries +0.9} \\
    \bottomrule
    \end{tabular}
\end{table}

FTP inherits FTW's geographic coverage and further filters it by PlanetScope availability, so it over-represents Europe and under-represents persistently cloudy and sparsely labeled regions (\Cref{sec:limitations}). Models evaluated and conclusions drawn on FTP may transfer unevenly to under-represented smallholder geographies. The country-level retention tables and patch-level quality index show which samples were kept. These records let users filter samples or reweight them when evaluating models.

\section{Held-out post-processing sweep}
\label{app:heldout_sweep}

The sweep varies augmentation, watershed, D4 test-time augmentation (TTA), auxiliary heads, and backbone size and reports object F1 under the released PRUE pixel-instance procedure rather than vectorized-polygon RQ. Results are macro-averaged over the ten dense-label held-out countries, excluding presence-only Kenya and retaining Portugal.

\paragraph{Augmentation settings.}
PRUE+ is implemented with Kornia~\cite{riba2020kornia}. It retains the released PRUE preprocessing: {\small\texttt{preprocess\_aug}} samples a divisor from $[5000,15000]$, and {\small\texttt{RandomResizedCrop}} uses scale $0.3$--$0.9$, ratio $0.75$--$1.33$, and $p=0.5$. PRUE+ then adds seasonal-window swap and per-band gamma jitter ($\gamma\sim\mathcal{U}[0.8,1.2]$, $p=0.3$).

The geometry/noise recipe adds affine intensity jitter ($a\sim\mathcal{U}[0.9,1.1]$, $b\sim\mathcal{U}[-0.02,0.02]$, $p=0.3$), Gaussian blur ($3\times3$, $\sigma\sim\mathcal{U}[0.5,1.5]$, $p=0.3$), Gaussian noise ($\sigma=0.015$, $p=0.3$), single-window dropout ($p=0.15$), rotation ($\pm30^\circ$, $p=0.5$), shear ($\pm5^\circ$, $p=0.3$), and boundary-thickness jitter by random binary dilation of $0$--$2$ pixels ($p=0.5$).

\end{document}